\documentclass[runningheads]{llncs}
\usepackage{graphicx}

\usepackage{tikz}
\usepackage{comment}
\usepackage{amsmath,amssymb} 
\usepackage{color}
\usepackage{cite}

\usepackage[accsupp]{axessibility}  

\usepackage[width=122mm,left=12mm,paperwidth=146mm,height=193mm,top=12mm,paperheight=217mm]{geometry}

\usepackage{overpic}
\usepackage{booktabs}
\usepackage{multirow}
\usepackage{rotating}
\usepackage{wrapfig}
\usepackage{amsmath}
\usepackage{amssymb}
\usepackage{bbm}
\usepackage{dsfont}
\usepackage{wrapfig}
\usepackage{verbatim} 
\usepackage{nicefrac} 

\newcommand{\tabref}[1]{Tab. \ref{#1}}
\newcommand{\figref}[1]{Fig. \ref{#1}}

\def\ie{\emph{i.e.}}
\def\eg{\emph{e.g.}}
\def\etc{\emph{etc}}
\def\etal{{\em et al.~}}
\def\sArt{{state-of-the-art~}}

\newcommand{\mathvec}[1]{\boldsymbol{#1}}
\newcommand{\mathmat}[1]{\mathbf{#1}}

\begin{document}
\pagestyle{headings}
\mainmatter
\def\ECCVSubNumber{3023}  

\title{Image Harmonization by \\ Matching Regional References} 
\authorrunning{Zhu et al.}
\author{Ziyue Zhu \quad Zhao Zhang \quad Zheng Lin \quad Ruiqi Wu  \quad Chunle Guo}
\institute{TKLNDST, CS, Nankai University \\
  \email{zhuziyue@mail.nankai.edu.cn; guochunle@nankai.edu.cn}
}

\maketitle

\begin{abstract}
  To achieve visual consistency in composite images, recent image harmonization methods typically summarize the appearance pattern of global background and apply it to the global foreground without location discrepancy.
  However, for a real image, the appearances (illumination, color temperature, saturation, hue,
  texture, \etc) of different regions can vary significantly.
  So previous methods, which transfer the appearance globally, are not optimal.
  Trying to solve this issue, we firstly match the contents between the foreground and background and then adaptively adjust every foreground location according to the appearance of its content-related background regions.
  Further, we design a residual reconstruction strategy, that uses the predicted residual to adjust the appearance, and the composite foreground to reserve the image details.   
  Extensive experiments demonstrate the effectiveness of our method. 
  The source code will be available publicly.
  \keywords{Image Harmonization, Regional Matching}
\end{abstract}

\section{Introduction}
Image harmonization aims at achieving visual consistency within the composite images.
When pasting the foreground of one image onto the background of another image,
the appearance (illumination, color temperature, saturation, hue, texture, \etc{}) of the composite foreground is typically
inconsistent with the background appearance.
To solve this problem, image harmonization methods adjust foreground appearance according to the background environment.
%

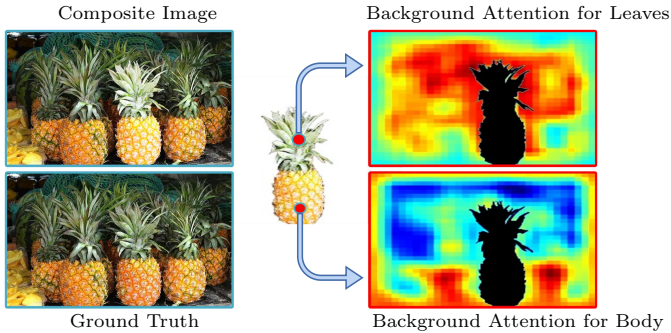
\begin{figure}[t]
  \vspace{15pt}
  \centering
  \begin{overpic}[width=0.65\linewidth]{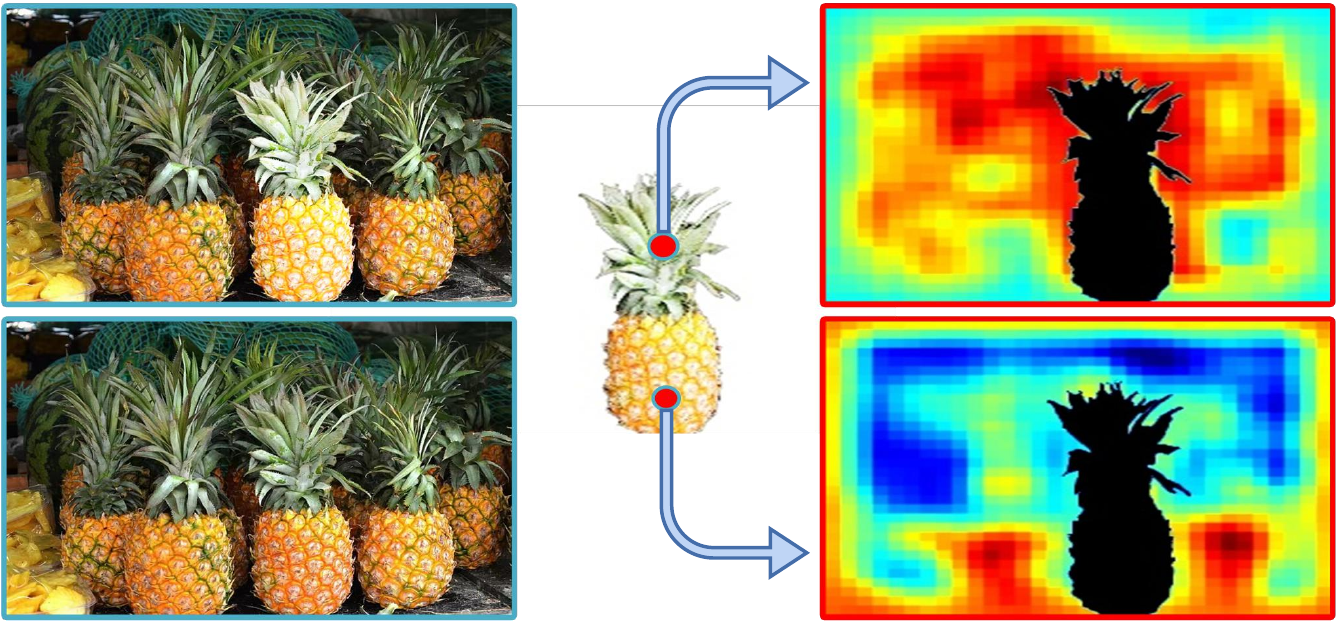}
  \put(9, 48){\scriptsize{Composite Image}}
  \put(11, -3){\scriptsize{Ground Truth}}
  \put(62, -3){\scriptsize{Background Attention for Body}}
  \put(61, 48){\scriptsize{Background Attention for Leaves}}
  \end{overpic}
  \caption{
  To adjust the disharmonious green leaves,
  a good reference is the other green leaves in background.
  The same is true for the disharmonious pineapple body.
  This observation motivates us to independently adjust different positions in the foreground according to its most referential background regions.
  As shown in the second column,
  with our method,
  more reasonable areas have made more contributions to harmonize the red positions.
  }
  \label{fig:intro}
  \vspace{-15pt}
\end{figure}

To obtain visual-consistent composite images, most learning-based methods adjust the appearance of the overall foreground according
to the visual information of the overall background by domain verification discriminator \cite{Cong2020DoveNet}, style transfer \cite{ling2021region}, reflectance and illumination \cite{Guo2021Intrinsic, Guo2021DHT}, channel attention \cite{cun2020improving, hao2020image}, and so on.
However, current methods ignore that, for a real image, the appearance of different regions can vary significantly.
Taking \figref{fig:intro} as an example, we can easily distinguish that the foreground pineapple is composite because its leaves and body are different from nearby ones.
However, the leaves and pineapple bodies share different appearance properties. 
In this case, a global background-to-foreground appearance translation strategy is not appropriate.

Based on the observation above, we think that a better solution to adjust the leaves is referring to the leaves in the background, so is the foreground pineapple body.
That is,
for each foreground location,
the content-related background regions may provide more reference cues.
In this paper,
we bring image harmonization more fine-grained adjusting,
where the appearance of each foreground location is determined by its related background regions.
Specifically,
we align the representation of each foreground location in appearance with matched background ones in a learned deep neural network (DNN) feature space.


Considering the different properties of the shallow and deep layers in DNN,
we design two appearance translation strategies to harmonize the foreground from coarse to fine.
1) For global adjustment,
we resort to deep but low resolution layers, whose pixel-level embedding fuses regional visual information and preserves the structure semantic well \cite{liu2020rethinking}.
We reconstruct each foreground location by linearly combining it with related background embeddings;
2) For more detailed adjustment,
we draw support from the shallow but high-resolution features,
which preserve more fine local content and appearance information.
Considering the significant appearance variance in different background regions, we capture diverse appearance information by cutting the background into overlapping patches.
And the appearance information of each patch is represented by its feature statistics.
For adaptively adjusting the appearance of each foreground location, 
we weighted transfer the appearance of the background patches to this location according to content similarity.
The similarity is measured by the correlation between foreground and background  content embeddings generated in a normalization manner.
In this way,
the visual appearance of each inconsistent foreground location is jointly determined by its related background patches.



Besides, in image harmonization,
the basic structure and texture should remain unchanged, and only the appearances, such as illumination and tone, are changed.
In this case, instead of directly generating the appropriate foreground \cite{ling2021region, Guo2021Intrinsic, Cong2020DoveNet, cun2020improving, BargainNet2021}, we retain the original composite input to keep the details while adjusting the visual appearances by our predicted residual. 
Without additional parameters and computation cost, it brings performance improvement to the model.
In summary, our major contributions are given as follows:
\begin{itemize} 
    \item We propose to adjust the appearance of each foreground location according to the related background. 
    Two appearance translation strategies are designed to achieve visual consistency in a coarse-to-fine manner.
    \item Focusing on learning the appearance change, we propose a reconstruction strategy, that adjusts foreground appearance by the predicted residual without losing the details of the original foreground.
    
    \item Our method achieves \sArt performance on the benchmark dataset \textit{iharmony4} \cite{Cong2020DoveNet}. Extensive ablation studies and visualizations validate the effectiveness of the proposed strategies.
\end{itemize}





 
\label{sec:related-work}
\section{Related Work}
\noindent\textbf{Image Harmonization} aims at  adjusting the appearance of the composite image foreground to be compatible with the background.
Traditional methods focus on matching low-level appearances based on statistics, such as global and local color distribution \cite{pitie2005ndim, reinhard2001color,pitie2007linear, xue2012understanding}, gradient information \cite{jia2006drag, perez2003poisson}, and multi-scale statistical features \cite{sunkavalli2010multi}. 
Recent methods further improve the realism of composite images with deep neural network.
For example, RainNet \cite{ling2021region} treats the mean and variance of deep representation as appearance information and adjusts the mean and variance of the foreground to be the same with that of the background.
DIH \cite{tsai2017deep} and Sofiiuk \etal \cite{sofiiuk2021foreground} utilize semantic information to capture the context of image, which helps harmonize the composite foreground.
DoveNet \cite{Cong2020DoveNet} designs a
domain verification discriminator, with the insight that the foreground needs to be transferred to the same domain as the background. 
BargainNet \cite{BargainNet2021} uses a domain code extractor to capture the background domain
information to guide the harmonization on the foreground.
S$^{2}$AM \cite{cun2020improving} extracts the global channel information of the background representation and weighted changes the foreground.
Hao \etal \cite{hao2020image} align the standard deviation of the foreground features with that of
the background features, capturing global dependencies in the entire image.
Guo \etal \cite{Guo2021Intrinsic, Guo2021DHT} disentangle composite image into reflectance and illumination for separate harmonization.

These learning-based methods regard that the composite foreground as a whole, 
and the overall appearance of the foreground is adjusted according to the overall background or part of the background.
Different from all existing methods, we start from the perspective that every location of the composite foreground should employ related background locations as reference to adjust its appearance.

\noindent\textbf{Style Transfer} is designed for changing the image style according to given style patterns while preserving content structure.
AdaIN \cite{huang2017arbitrary} adaptively shifts and re-scales the content representation so that it keeps the same mean and standard deviation as the style representation.
It inspired many style transfer methods \cite{wang2021rethinking, lin2021drafting, cheng2021style} and the works of other vision tasks \cite{huang2018multimodal, li2020advancing, liu2019few}.
WCT \cite{li2017universal} transfer the style by whitening the content representation and then coloring it with style representation.
SANet \cite{Park2019SANet} efficiently and flexibly integrates the local style patterns according to the semantic spatial distribution of the content image.
Similar to style transfer, image harmonization achieves visual consistency by transferring the background appearance to the foreground.
Recent method RainNet \cite{ling2021region} also validates that style transfer method AdaIN \cite{huang2017arbitrary} is very effective in image harmonization.



\section{Proposed Method}

\vspace{-20pt}
\begin{figure}[h!]
  \centering
  \begin{overpic}[width=0.9\linewidth]{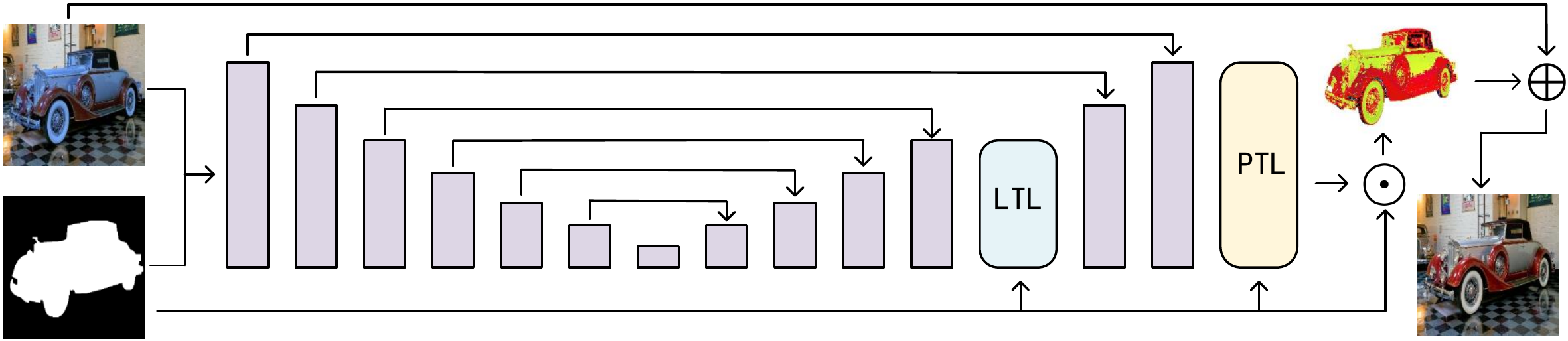}
  \end{overpic}
  \vspace{-10pt}
  \caption{
    \textbf{The overall framework of our method.}
    We employ the U-net structure to harmonize the composite foreground.  
    Two appearance translation modules (LTL and PTL) are designed to adjust the foreground appearance in a coarse to fine manner
  }
  \label{fig:m}
  \vspace{-10pt}
\end{figure}
\vspace{-20pt}
\subsection{Overview}
Given a pair of composite image $\mathmat{I}$ and its corresponding real image $\mathmat{H}$, 
image harmonization aims at harmonizing the composite foreground area, 
which is indicated by a binary mask $\mathmat{M}$.
Meanwhile the background mask is denoted as $\Bar{\mathmat{M}} = 1 - \mathmat{M}$.
Following \cite{Cong2020DoveNet, ling2021region, BargainNet2021, cun2020improving, Guo2021Intrinsic},
we employ the U-net structure as our Generator $\mathmat{G}$ to harmonize the foreground.

Considering the significantly variant appearances of different regions of the foreground and the background, applying the global appearance pattern of the background to the global foreground \cite{ling2021region, cun2020improving, Cong2020DoveNet, BargainNet2021, sofiiuk2021foreground} is probably not an optimal strategy.
We adjust the appearance of each composite foreground location by matching it with related background locations,
then transfer the appearance of matched background regions to the foreground location.
Following our idea of adjusting the foreground from coarse to fine, we design two appearance translation strategies on deep and shallow features respectively. 

Last but not least, we propose a simple but effective reconstruction strategy, which makes our generator $\mathmat{G}$ predict the residual for adjusting the foreground appearance rather than directly generate the foreground.
This strategy brings performance improvement without additional parameters and computation cost.

\begin{figure}[t!]
  \centering
  \begin{overpic}[width=0.95\linewidth]{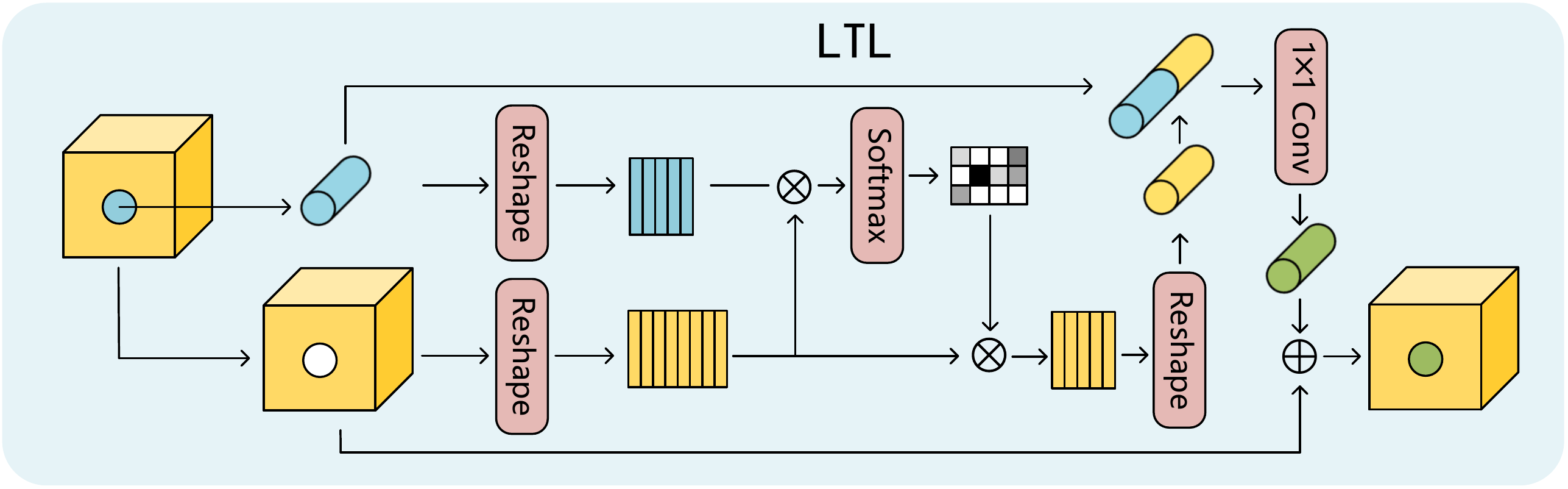}
  \end{overpic}
  \vspace{-10pt}
  \caption{
    \textbf{Locations-to-Location Translation (LTL).}
    In the deep layers, we fuse each foreground location with related background locations to achieve visual consistency.
  }
  \label{fig:m}
  \vspace{-10pt}
\end{figure}

\subsection{Locations-to-Location Translation (LTL)}
For deep features $\mathmat{F}\in\mathbb{R}^{C\times H\times W} $ with low resolution, the visual information is fused, and the structure semantic is well preserved \cite{liu2020rethinking}.
We adjust the foreground appearance coarsely on low-resolution features by fusing each foreground location with its related background locations.
Specifically, we employ an attention mechanism and regard the foreground locations in deep features as query to index related
background locations. 
The indexed background locations are linearly combined with the foreground so that the appearance of each foreground location is adaptively adjusted by its related background locations. 

We illustrate our strategy (LTL) in \figref{fig:m}.
After a self-attention layer and instance normalization, we separately extract the foreground and background parts of feature $\mathmat{F}$ and reshape them into foreground tokens $\mathmat{T}_f\in\mathbb{R}^{L_f\times C}$ and background tokens $\mathmat{T}_b\in\mathbb{R}^{L_b\times C}$, where $L_f + L_b = HW$.
Then we commit to adjusting the appearance of every foreground location according to its related background locations.
The related background tokens (locations) $\mathmat{T}_r$ are weighted indexed based on the similarity map between the foreground tokens $\mathmat{T}_f$ and background tokens $\mathmat{T}_b$,
\begin{equation}
\mathmat{T}_r = \text{Softmax}\left(\mathmat{T}_f  \mathmat{T}^T_b\right) \mathmat{T}_b \in\mathbb{R}^{L_f\times C}.
\end{equation}
Further, to adaptively adjust the appearance, we concatenate foreground tokens with the related background tokens and fuse them together with a linear layer. 
\begin{equation}
\mathmat{T}_h = \text{Linear}\left(\text{Concat}\left(\mathmat{T}_f,  \mathmat{T}_r\right)\right) \in\mathbb{R}^{L_f\times C}.
\end{equation}
Then, the fused tokens $\mathmat{T}_r$ and the background tokens are reshaped back to $\mathmat{F}_h$ for further steps.

Our LTL fuses every foreground location in deep features with the related background locations.
Under ground truth supervision, the appearance of the background is adaptively transferred to the foreground.
Although previous work \cite{hao2020image} also employs an attention map, it actually applies channel attention on the global foreground but we independently adjust each foreground location.


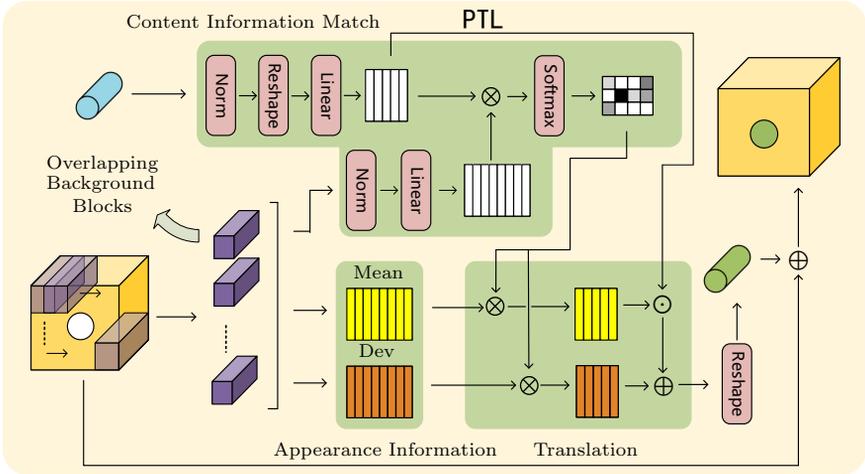
\begin{figure}[t!]
  \centering
  \begin{overpic}[width=0.95\linewidth]{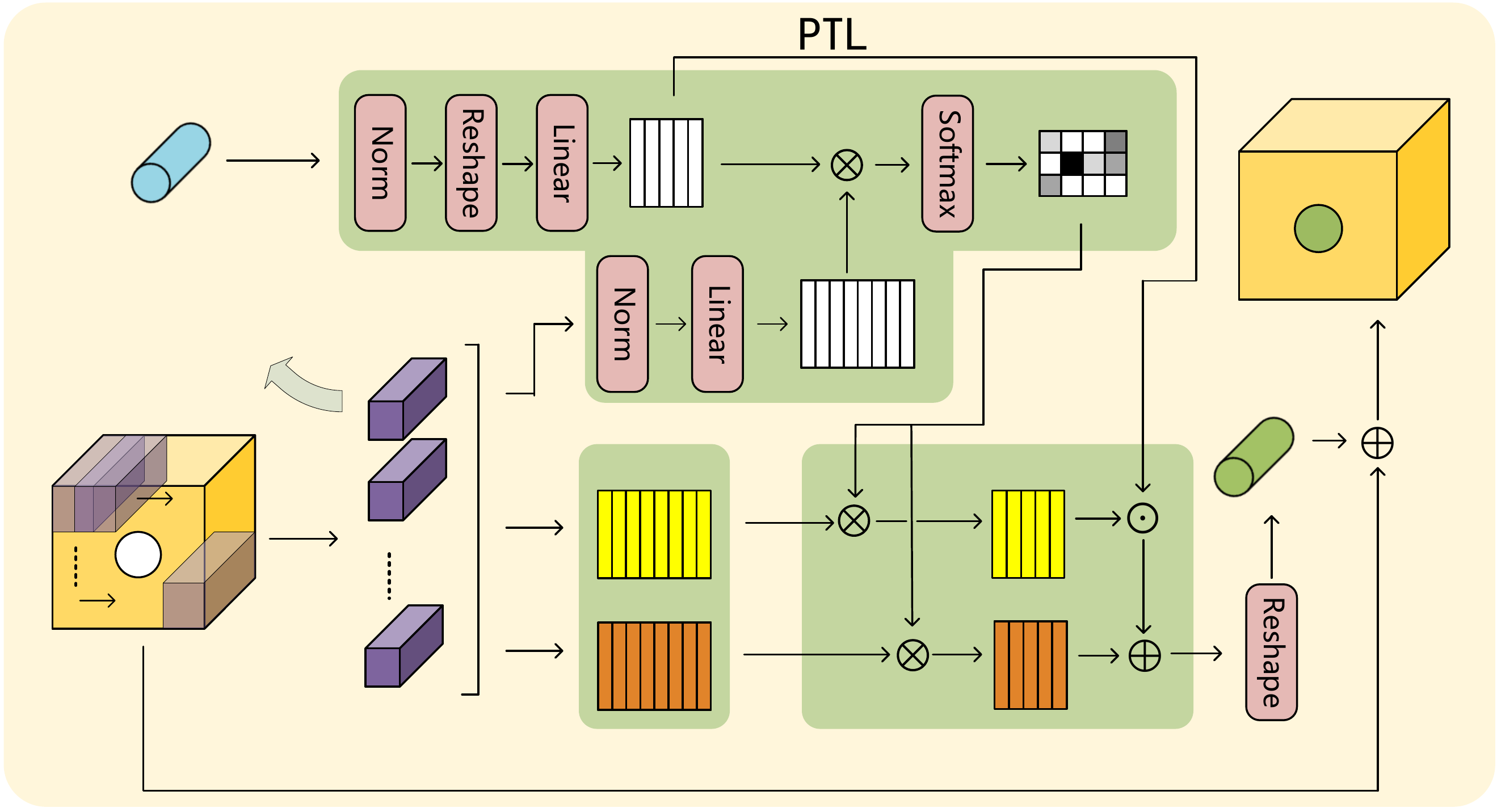}
  \put(31.5, 2.5){\scriptsize{Appearance Information}}
  \put(61.5, 2.5){\scriptsize{Translation}}
  \put(14.5, 51){\scriptsize{Content Information Match}}
  \put(40.7, 22.6){\scriptsize{Mean}}
  \put(41.3, 13.7){\scriptsize{Dev}}
  \put(5.3, 35){\scriptsize{Overlapping}}
  \put(5.3, 32.7){\scriptsize{Background}}
  \put(8.3, 30.1){\scriptsize{Blocks}}
  \end{overpic}
  \caption{
    \textbf{Patches-to-Location Translation (PTL).} 
    We decompose the high resolution image features,
    which contain rich visual cues,
    into content information and appearance information.
    The foreground appearances are changed according to the similarity of the content information.
  }
  \vspace{-10pt}
  \label{fig:PTL}
\end{figure}
\vspace{-10pt}

\subsection{Patches-to-Location Translation (PTL)}
To finely grained adjust the foreground, we propose an appearance translation strategy on high resolution features, which are more related to visual-consistency information (\eg{}, color tone and illumination) \cite{ling2021region}. 
Specifically, we match the content information between the foreground locations and background regions, which are overlapping background patches in our method.
Then, the appearances of different background regions are weighted transferred to each foreground location based on the similarity between their contents.

The way we extract the content and appearance information is inspired by AdaIN \cite{huang2017arbitrary}, proving that instance normalization can discard the original appearance information while remaining content information.
Meanwhile, the appearance information can be represented by the statistics (mean and standard deviation) of the image feature.
Recent method RainNet \cite{ling2021region} has validated that transferring the overall background appearance to the overall foreground by AdaIN \cite{huang2017arbitrary} is very effective in image harmonization.
However, the appearances of different background regions can vary significantly, and such a global-to-global style translation strategy is not quite reasonable.

Back to our method, to fully capture both appearance and content information of different background regions, we cut the background features $\mathmat{F}\times \Bar{\mathmat{M}} $ into multiple blocks $\{\mathmat{B}_n\}_{n=1}^{K}$ with the stride of $(\frac{H}{32}, \frac{W}{32})$ and the block size of $C\times \frac{H}{8} \times \frac{W}{8}$.
Besides, the background mask $\mathmat{\Bar{M}}$ is also divided into corresponding patches $\{\mathmat{P}_n\}_{n=1}^{K}$.
The appearance of each block can be revealed by its statistics, while the block after normalization contains the content information \cite{huang2017arbitrary}.
Specifically, we use the mean $\mu_{n} \in\mathbb{R}^{1\times C}$ and standard deviation $\sigma_{n} \in\mathbb{R}^{1\times C}$ of each block to represent its appearance,
\begin{footnotesize}
\begin{equation}
\begin{aligned}
\mu_{n} &= \frac{1}{\sum\limits_{h, w} \mathmat{P}_{n}^{h,w} } \sum\limits_{h, w} \mathmat{B}_{n}^{h,w} \times \mathmat{P}_{n}^{h,w} \\
\sigma_{n} &= \sqrt{\frac{1}{\sum\limits_{h, w} \mathmat{P}_{n}^{h,w} } (\sum\limits_{h, w} \mathmat{B}_{n}^{h,w} \times \mathmat{P}_{n}^{h,w}-\mu_{n})^2 + \epsilon},
\end{aligned}
\label{eq:stat}
\end{equation}
\end{footnotesize}
where $h,w$ is a location within the background patch
and ``$\times$'' denotes element-wise multiplication.

For subsequent steps, we compile the appearance of each background block (represented by $\mu_n$ and $\sigma_n$) into $\mu\in\mathbb{R}^{K\times C}$ and $\sigma\in\mathbb{R}^{K\times C}$.
To extract the foreground and background content information, we apply instance normalization (IN) \cite{ulyanov2016instance} on each background block $\mathmat{B}_n$ as well as the foreground region  $\mathmat{F}\times \mathmat{M} $ independently. 
For matching the foreground content with the background, we squeeze  background blocks $\{\mathmat{B}_n\}_{n=1}^{K}$ after IN into background content tokens $\{\mathvec{b}_n\in\mathbb{R}^{1\times C}\}_{n=1}^{K}$ with a linear projection.
After independent IN within the foreground region, the foreground locations in feature $\mathmat{F}$ are reshaped into foreground content tokens $\{\mathvec{f}_n\in\mathbb{R}^{1\times C}\}_{n=1}^{L_f}$.
So far, we have obtained the foreground content $\mathmat{C}_f\in\mathbb{R}^{L_f\times C}$ as well as the background content of multiple blocks $\mathmat{C}_b\in\mathbb{R}^{K\times C}$ with the corresponding appearance tokens $\mu\in\mathbb{R}^{K\times C}$,
$\sigma\in\mathbb{R}^{K\times C}$.


Then we transfer the appearance of background blocks to foreground locations based on the similarity between their contents.
Concretely, the attention map between the content of foreground locations and background blocks is calculated to measure their correlation.
Then the attention map is separately applied on the mean $\mu$ and standard deviation $\sigma$ to weighted index the background appearance tokens for each foreground location.
\begin{equation}
\begin{aligned}
\mathmat{A} &= \text{Softmax}\left(\mathmat{C}_f  \mathmat{C}^T_b\right)\mu
\in\mathbb{R}^{L_f\times C} \\
\mathmat{V} &= \text{Softmax}\left(\mathmat{C}_f  \mathmat{C}^T_b\right)\sigma\in\mathbb{R}^{L_f\times C}.
\end{aligned}
\label{eq:matching}
\end{equation}
$\mathmat{A}$ and $\mathmat{V}$ represent the indexed appearance for all foreground locations.
We transfer the appearance to the foreground content $\mathmat{C}_{f}$ with the following equation.
\begin{equation}
\mathmat{T}_{fb} = \mathmat{C}_{f} \times \mathmat{V} + \mathmat{A}.
\label{eq:translation}
\end{equation}
Finally, the foreground after translation $\mathmat{T}_{fb}$ are reshaped back to the original location.

  
  

  
    

    
    
  
    
    


Different from the existing image harmonization method \cite{ling2021region}, which shifts and scales the overall foreground region with the mean and standard deviation of the overall background, we adaptively adjust each foreground location with the appearance of its content-related background regions.

\begin{figure}[h!]
  \centering
  \begin{overpic}[width=0.95\linewidth]{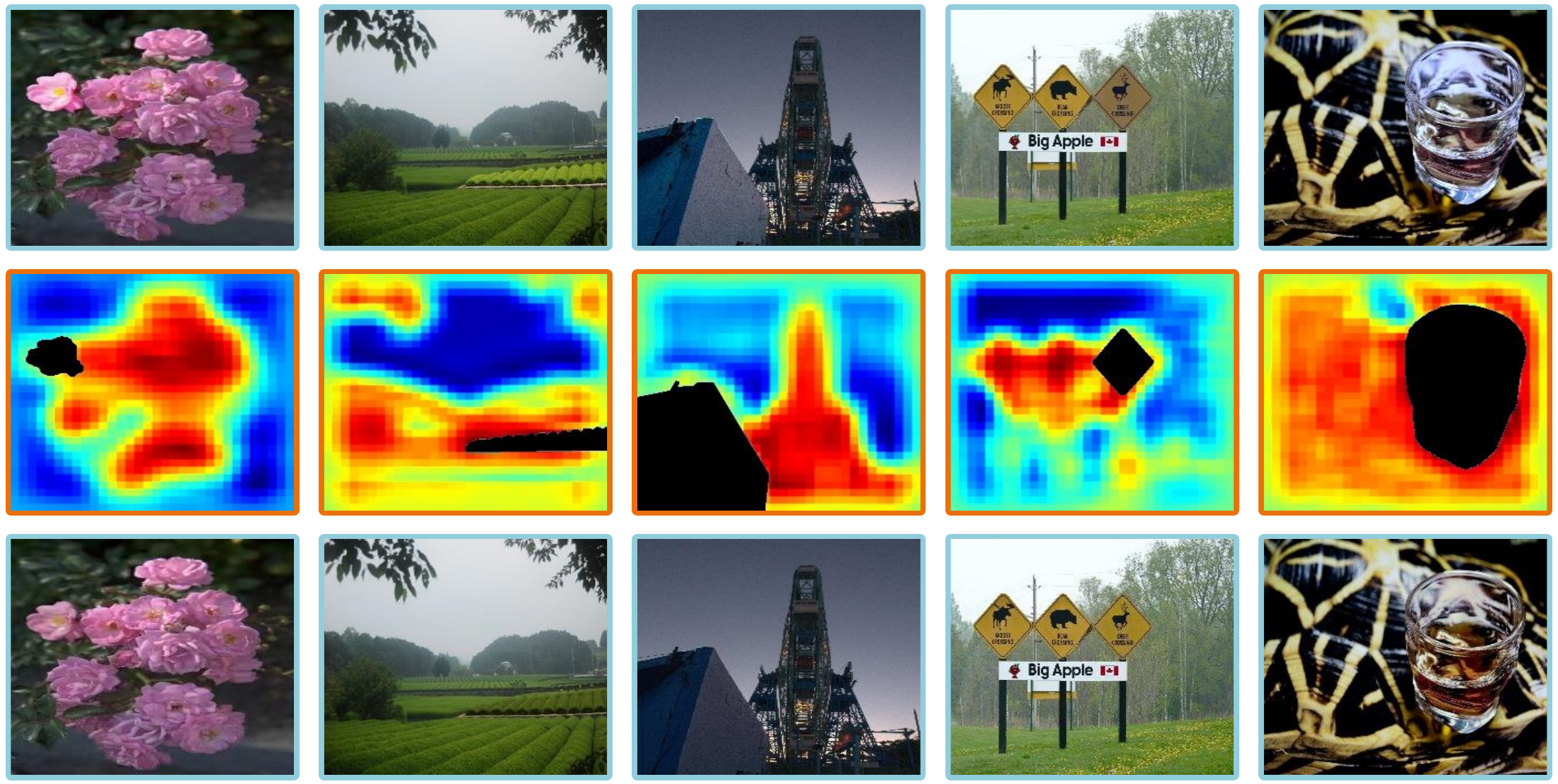}
  \end{overpic}
  \caption{
    \textbf{Visualization of contribution map in PTL.}
    When harmonizing the first-row composite images into the third-row predictions,
    we illustrate the background regions, where the overall foregrounds pay high attention to in the second row.
  }
  \label{fig:hot}
  \vspace{-10pt}
\end{figure}

In \figref{fig:hot}, we illustrate the attention maps for the overall foreground and our predictions.
Each foreground of the composite images in the first three columns has related background regions, while the foreground of the last image does not have such regions.
We can see that the first three composite foregrounds (plants, square building, and the warning sign) successfully find their content-related backgrounds.
The third row shows that the three foregrounds are harmonized by the background regions with high attention.
When not containing clearly relevant background regions, the composite glass in the last column pays attention to the global background.
In this case, the foreground appearance is changed by the global background.

\subsection{Residual Reconstruction Layer}


Image harmonization aims at adjusting the appearance of the composite foreground while preserving the basic structure and content information.
In other words, the adjustment is relatively slight, and the model only needs to learn the change of appearance.
Given a composite image $\mathmat{I}$ and its foreground mask $\mathmat{M}$, previous works \cite{Cong2020DoveNet, Guo2021Intrinsic, BargainNet2021, liu2020rethinking, cun2020improving} typically employ a generator to directly predict the foreground $\hat{\mathmat{I}}\times \mathmat{M}$.
Combining with the real background, the final output $\mathmat{O}$ is generated as follow:
\begin{equation}
\mathmat{O} = \mathmat{I} \times (1-\mathmat{M}) + \hat{\mathmat{I}} \times \mathmat{M}
\end{equation}
However, such strategy requires the model to generate a harmonious appearance of the foreground while preserving the original details at the same time.


From our perspective, image harmonization model should focus on learning the appearance change and rely on the composite input to preserve the details.
Instead of generating the foreground directly, we aim at predicting the residual for changing the appearance.
\begin{equation}
\mathmat{O} = \mathmat{I} + \hat{\mathmat{I}} \times \mathmat{M}
\end{equation}
Our strategy improves our model performance without additional parameters and computation cost. 
The effectiveness will be further validated in the ablation study.



\section{Experiments}

\subsection{Implementation Details}
\noindent\textbf{Training Details.} Our model is trained for 140 epochs with Adam optimizer with $\beta$ = 0.9, $\beta$ = 0.999 and initial learning rate of 0.001. 
The learning rate is divided by 10 at 120th epoch.
We resize input images as $256\times256$ during both training and testing.
The training data is augmented by random crop and random horizontally flip.
All the experiments are implemented in Pytorch \cite{pytorch2019paszke} with the batch size of 4 on an Nvidia 2080Ti GPU.

\noindent\textbf{Loss Function.}
Following \cite{sofiiuk2021foreground}, which takes the area of foreground region into account, we employ the foreground MSE loss as our loss function,
\begin{equation}
\mathcal{L}(\hat{\mathmat{I}}, \mathmat{I}) = \frac{\sum\limits_{h, w}\left \|\hat{\mathmat{I}}_{h,w}-\mathmat{I}_{h,w}\right\|_2^2}{\max\{A_{min}, \sum\limits_{h, w}\mathmat{M}_{h,w}\}},
\end{equation}
$A_{min}$ is a hyperparameter for preventing instability during training, and we set it into 100 as suggested in \cite{sofiiuk2021foreground}.

\subsection{Datasets and Evaluation Metrics}
\noindent\textbf{Datasets.} 
Same with recent works \cite{Cong2020DoveNet, BargainNet2021, Guo2021Intrinsic, sofiiuk2021foreground, ling2021region}, 
we conduct the experiments on the benchmark dataset \textit{iHarmony4} consisting of 4 sub-datasets (\ie{HCOCO, HAdobe5K, HFlicker, and Hday2night}), 
and 73147 pairs of synthesized composite images and corresponding ground truth images.
We also follow the train-split suggestion \cite{Cong2020DoveNet} and each
sub-dataset is split into training and test sets.

\noindent\textbf{Evaluation Metrics.} Following \cite{sofiiuk2021foreground}, we use mean squared error (MSE), foreground mean squared error (fMSE), and Peak Signal-to-Noise Ratio (PSNR) to evaluate our model performance.
Note that fMSE computes the MSE of the foreground.
Image harmonization only changes the composite foreground, but the ratios of the foregrounds vary greatly across images.
Compared with MSE, fMSE is not affected by the ratio and can better reveal the performance.

\subsection{Comparison with Current Methods}
To illustrate performance, we qualitatively and quantitatively compare two versions of our method with 6 \sArt methods, including DoveNet \cite{Cong2020DoveNet}, S$^2$AM \cite{cun2020improving}, BargainNet \cite{BargainNet2021}, Intrinsic \cite{Guo2021Intrinsic}, RainNet \cite{ling2021region}, and D-HT \cite{Guo2021DHT}.

\begin{table}[t!]
\scriptsize
\renewcommand{\arraystretch}{1.5}
\renewcommand{\tabcolsep}{1mm}
\begin{tabular}{l|l|cccccc|cc}
\hline\toprule
     & & DoveNet  & S$^2$AM & BargainNet &  Intrinsic & RainNet  & D-HT  &   &  \\ 
    [-0.2cm]
   &  & \tiny CVPR20  & \tiny TIP20 & \tiny ICME21  & \tiny CVPR21 & \tiny CVPR21   & \tiny ICCV21  &  Ours  &  Ours-p \\ 
   [-0.2cm]
    Dataset & Metric  & \cite{Cong2020DoveNet}  &  \cite{cun2020improving} & \cite{BargainNet2021}  &  \cite{Guo2021Intrinsic} & \cite{ling2021region}  & \cite{Guo2021DHT}  &  \\
   \hline
           & PSNR$\uparrow$   & 35.83 & 35.47 & 37.03 & 37.16 & 37.08 & 38.76 & \underline{39.15} & \textbf{39.75}     \\
HCOCO      & MSE$\downarrow$    & 36.72 & 41.07 & 24.80 & 24.92 & 31.12 & 16.89 & \underline{16.13} & \textbf{13.63}     \\
           & fMSE$\downarrow$  & 551.01 & 542.06 & 397.52 & 416.38 & 535.39 & 299.30 & \underline{278.61} & \textbf{244.55}     \\ \hline
           & PSNR$\uparrow$   & 34.34 & 33.77 & 35.34 & 35.20 & 36.21 & 36.88 & \underline{37.34} & \textbf{38.04}     \\
HAdobe5k   & MSE$\downarrow$   & 52.32  & 63.40 & 39.92 & 43.02 & 42.85 & 38.53 & \underline{25.41} & \textbf{20.83}     \\
           & fMSE$\downarrow$  & 380.39 & 404.62 & 279.38 & 284.21 & 320.43 & 265.11 & \underline{194.48} & \textbf{168.08}     \\ \hline
           & PSNR$\uparrow$   & 30.21 & 30.03 & 31.34 & 31.34 & 31.64 & 33.13 & \underline{33.51} & \textbf{34.06}     \\
HFlickr     & MSE$\downarrow$   & 133.14 & 143.45 & 97.32 & 105.13 & 117.59 & 74.51 & \underline{67.74} & \textbf{59.08}     \\
           & fMSE$\downarrow$  & 827.03 & 785.65 & 700.19 & 716.60 & 751.12 & 515.45 & \underline{469.23} & \textbf{412.19}    \\ \hline
           & PSNR$\uparrow$  & 35.27 & 34.50 & 35.69 & 35.96 & 34.83 & 37.10 & \textbf{37.62} & \underline{37.61}     \\
Hday2night & MSE$\downarrow$  & 51.95 & 76.61 & 50.87 & 55.53 & 47.24 & 53.01 & \textbf{46.79} & \underline{47.96}     \\
           & fMSE$\downarrow$  & 1075.71 & 989.07 & 835.06 & 797.04 & 852.08 & 704.42 & \underline{685.36} & \textbf{641.79}     \\ \hline
           & PSNR$\uparrow$  & 34.76  & 34.35 & 35.87 & 35.90 & 36.12 & 37.55 & \underline{37.97} & \textbf{38.57}     \\
All        & MSE$\downarrow$  & 52.33  & 59.67 & 37.79 & 38.71 & 40.29 & 30.30 & \underline{25.16} & \textbf{21.43}     \\ 
           & fMSE$\downarrow$  & 532.62 & 594.67 & 404.76 & 400.29 & 469.60 & 320.78 & \underline{282.69} & \textbf{248.12}     \\ \bottomrule
\end{tabular}
\vspace{10pt}
\caption{
		\textbf{Quantitative comparisons} of PSNR, MSE and fMSE on the benchmark dataset \textit{iharmony4} \cite{Cong2020DoveNet} by two versions of our methods and other \sArt methods. 
		``$\uparrow$'' means that the higher the numerical value, the better the model performance.
		The numerical value with an underline refers to the second hightest value.
		Specifically, the difference between our two versions of the method lies in the amount of parameters, which will be discussed in detail in our ablation study.
	}
    \label{tab:quantitative} 
\end{table}

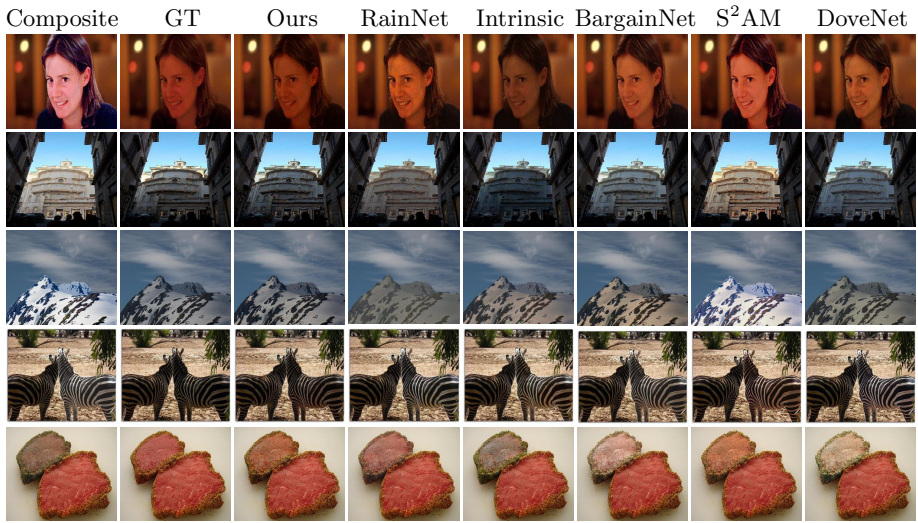
\begin{figure*}[h!]
\vspace{10pt}
  \centering
  \begin{overpic}[width=1.00\textwidth]{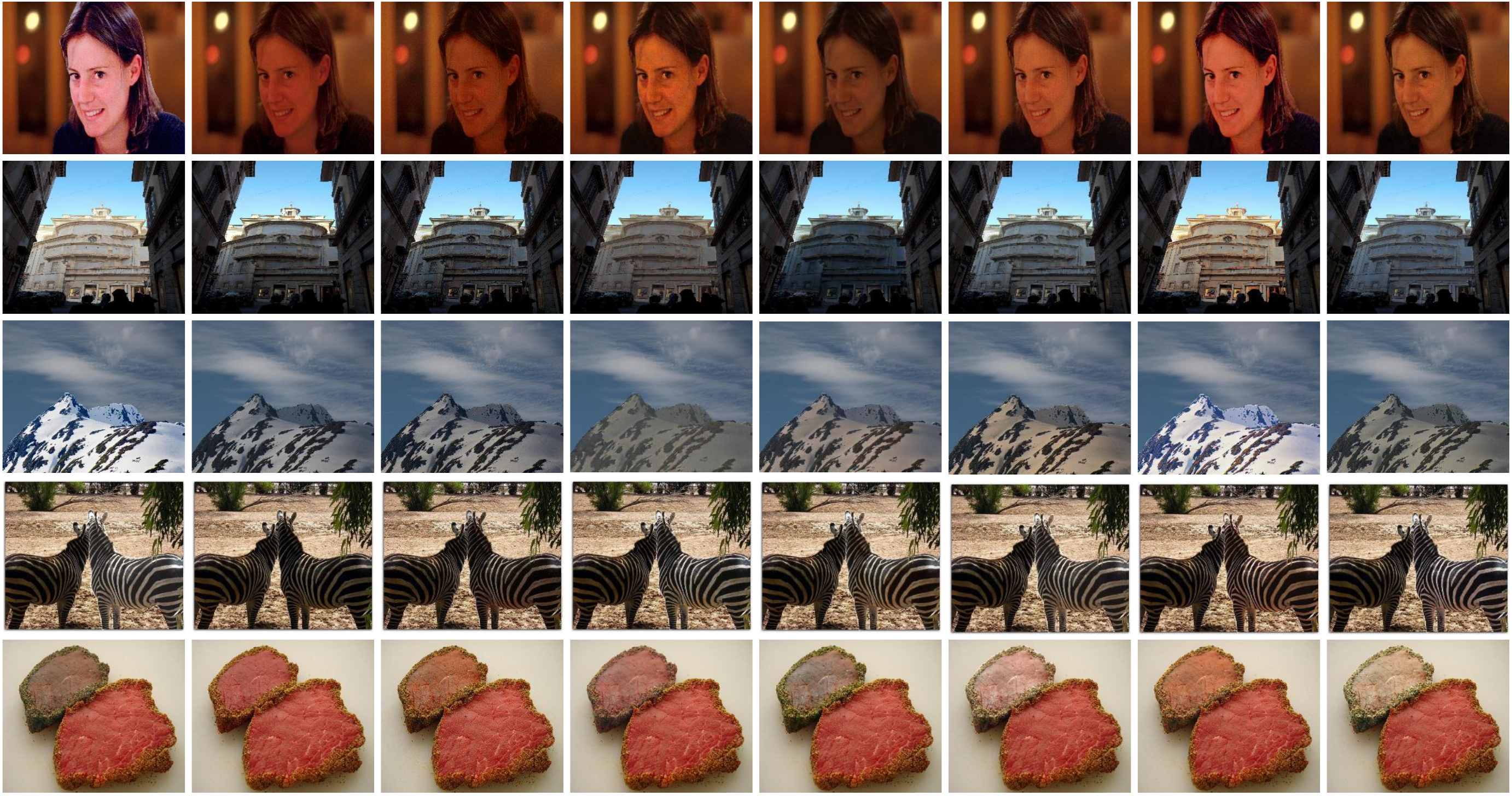}
    \put(0.1, 53.5){Composite}
    \put(17.5, 53.5){GT}
    \put(28.7, 53.5){Ours} 
    \put(39.0, 53.5){RainNet }
    \put(51.7, 53.5){Intrinsic}
    \put(62.5, 53.5){BargainNet }
    \put(78, 53.5){S$^2$AM } 
    \put(89, 53.5){DoveNet } 
  \end{overpic}
  \caption{
		\textbf{Visual comparison} of our method with five \sArt methods. 
	 }
  \label{fig:compare}
\end{figure*}

\noindent\textbf{Quantitative Comparison.}
In \tabref{tab:quantitative}, we compare two versions of our methods with other \sArt image harmonization methods in quantitative results.
The difference between our two versions of the method lies in the parameter number, which will be discussed in detail.
We can see that our method outperforms all of them across all sub-datasets and all metrics.
Compared with the most recent method D-HT \cite{Cong2020DoveNet} on \textit{iHarmony4} dataset, our method brings 5.14 improvement in terms of MSE, 38.01 improvement in terms
of fMSE and 0.42 dB improvement in terms of PSNR.
When we increase the channel number of our model, the improvements in terms of MSE, fMSE and PSNR increase to 8.87, 72.66 and 1.02 dB.
Our performance on the benchmark dataset fully demonstrates the effectiveness of our method.

\noindent\textbf{Qualitative Comparison.}
In \figref{fig:compare}, we compare the predictions of our methods with five other methods on the benchmark dataset \textit{iHarmony4} \cite{Cong2020DoveNet}.
Among the five groups of images. three of them have clearly related background regions,
while the other two do not. 
However, our method achieves high visual performance on all the composite images.
Without clear reference, our method still harmonizes the composite women and the mountain peak successfully.
Referring to background architecture, zebra and meat, our method achieves even better performance on the remaining three composite images.
Especially for the composite meat, disturbed by the white background, some methods make the meat too white.
In contrast, our method successfully adjusts the meat by the related background.

\section{Ablation Study}
In this section, we illustrate the effectiveness of locations-to-location translation (LTL), patches-to-location (PTL) as well as our residual reconstruction layer. 
To demonstrate the effectiveness of matching content-related background, we compare our PTL with AdaIN \cite{huang2017arbitrary}, which transfers the global background appearance to the global foreground.
Meanwhile, we also show how the size and stride of cutting background patches in our PTL influence the performance.
Besides, we compare the performance and parameters of our method with others.
Finally, to emphasize the influence of parameter size, we compare our model performance under different channel numbers.

\begin{table}[t!]
	\centering
	\renewcommand{\arraystretch}{1.25}
	\renewcommand{\tabcolsep}{2mm}
	\begin{tabular}{c|l|ccc}
		\hline\toprule
		\multirow{2}{*}{ID} & \multirow{2}{*}{Combination}   & \multicolumn{3}{c}{\textit{iHarmony4}~\cite{Cong2020DoveNet}}     \\
		       &  &  PSNR$\uparrow$ & MSE$\downarrow$ & fMSE$\downarrow$   \\
		\hline
		1  & Base                          
		& 37.27 & 30.84 & 330.88    \\
	    2  & RBase            
	    & 37.41 & 29.27 & 318.22    \\
		3  & RBase + LTL 
		& 37.66 &  27.00 & 307.38    \\
		4  & RBase + Self-Attention
		& 37.48 & 29.03 & 315.57   \\
		5  & RBase + AdaIN \cite{huang2017arbitrary}
		& 37.37 & 28.48 & 317.39    \\
		6  & RBase + PTL 
		& 37.70 & 27.40 & 299.70    \\
		 
	    7 & RBase + LTL + PTL  
		& 37.97 & 25.16 & 282.69    \\
		\hline
		 & RBase-p + LTL + PTL  
		& 38.57 & 21.43 & 248.12    \\
		\bottomrule
	\end{tabular}
	\vspace{10pt}
    \caption{
		\textbf{Ablation study} of our method on the overall benchmark dataset \textit{iharmony4} \cite{Cong2020DoveNet}.
		``Base'' denotes our baseline model.
		``RBase'' denotes our baseline with residual reconstruction layer.
		``LTL'' and ``PTL'' are our appearance translation strategies.
		 ``AdaIN'' denotes the RAIN module in RainNet \cite{ling2021region}.
		 ``RBase-p'' means increasing the channel number upon ``RBase''.
	}
	\label{tab:ABL}
	\vspace{-20pt}
\end{table}

\noindent\textbf{Effectiveness of Residual Reconstruction Layer.}
Following \cite{Guo2021Intrinsic, ling2021region}, we employ the U-net structure to directly predict the foreground as our baseline model.
In \tabref{tab:ABL}, we use ``RBase'' to denote the baseline model with residual reconstruction layer.
Compared with the previous reconstruction strategy, we predict the residual for adjusting the foreground appearance and bring performance improvement of 0.14dB, 1.57 and 12.66 in terms of PSNR, MSE and fMSE.
 
\noindent\textbf{Effectiveness of LTL.}

Our locations-to-location translation strategy (LTL) adjusts the appearance of each foreground location by related background locations on low-resolution features.
It can be freely added to different layers in our decoder.
In \tabref{tab:ABL}, we can see that adding LTL to our model brings 0.25dB, 2.27 and 10.84 performance improvement in terms of PSNR, MSE and fMSE.
Meanwhile, in \tabref{tab:Resolution} we add our LTL to the layers of our decoder with different resolutions.
We can see that adding our LTL to the decoder layer with higher resolution can achieve better performance.
Compared with adding LTL to the layer with $16 \times 16$ resolution, adding LTL to $64 \times 64$ resolution brings performance improvement of 0.13dB, 1.24 and 7.62 in terms of PSNR, MSE and fMSE.
Further, we also apply our LTL to the three layers.
Such multi-scale strategy can make our training process more stable and converge to better performance. 
However, with the increase of the feature resolutions,
the memory cost also increases dramatically.
Considering the balance between the performance and computation cost, we only add our LTL to one layer with $32\times32$ resolution in our final model.

\begin{table}[t!]
	\centering
	\renewcommand{\arraystretch}{1.25}
	\renewcommand{\tabcolsep}{1.5mm}
	\begin{tabular}{c|c|ccc}
		\hline\toprule
		\multirow{2}{*}{Resolution} & \multirow{2}{*}{Memory (MB)} &    \multicolumn{3}{c}{\textit{iHarmony4}~\cite{Cong2020DoveNet}}     \\
		        &  &  PSNR$\uparrow$ & MSE$\downarrow$ & fMSE$\downarrow$   \\
		\hline
                 
		16 $\times$ 16 & 97 & 37.44 &  28.91 & 319.31    \\
	                
	    32 $\times$ 32 & 120 & 37.57 & 28.38 & 312.06    \\
		    
		64 $\times$ 64 & 379 & 37.57 & 27.67 & 311.69    \\   \hline 
		            
		All & 410 & 37.66  &  27.00 & 307.38    \\
		\bottomrule
	\end{tabular}
	\vspace{6pt}
	\caption{
		\textbf{Performance of adding LTL to different layers.} We add our LTL to the decoder layers with different resolutions.
		``All'' denotes adding LTL to the three layers.}
	\label{tab:Resolution}
	\vspace{-30pt}
\end{table}

\noindent\textbf{Effectiveness of PTL.}
Our PTL adjusts the appearance of each foreground location by the content-related background patches.
\tabref{tab:ABL} demonstrates that our PTL brings 0.29dB, 1.87 and 18.52 performance improvement in terms of PSNR, MSE and fMSE. 
Note that we choose to cut our background into overlapping patches with the patch size of $(\frac{h}{8}, \frac{w}{8})$ and the stride of $(\frac{h}{32}, \frac{w}{32})$ in our final model.
In \tabref{tab:Patch} we cut the background into patches with  different patch sizes and stride sizes.
With a smaller stride size, the background are cut into larger number of patches.

\begin{table}[t!]
	\centering
	\renewcommand{\arraystretch}{1.25}
	\renewcommand{\tabcolsep}{1.2mm}
	\begin{tabular}{c|c|c|ccc}
		\hline\toprule
		\multirow{2}{*}{Patch Size} & \multirow{2}{*}{Stride Size} & \multirow{2}{*}{Memory (MB)}  & \multicolumn{3}{c}{\textit{iHarmony4}~\cite{Cong2020DoveNet}}     \\
		      & &  &  PSNR$\uparrow$ & MSE$\downarrow$ & fMSE$\downarrow$   \\
		\hline
        & $(\frac{h}{4}, \frac{w}{4})$           
		& 175 & 37.46 &  28.61 & 315.05    \\
	     $(\frac{h}{4}, \frac{w}{4})$  & $(\frac{h}{8}, \frac{w}{8})$            
	    & 214 & 37.57 & 28.65 & 311.76    \\
		   & $(\frac{h}{16}, \frac{w}{16})$ 
		& 348 & 37.61 & 27.30 & 305.71    \\   \hline 
		& $(\frac{h}{8}, \frac{w}{8})$           
		& 169 & 37.52 &  27.76 & 313.98    \\
	     $(\frac{h}{8}, \frac{w}{8})$  & $(\frac{h}{16}, \frac{w}{16})$            
	    & 230 & 37.62 & 27.33 & 302.55    \\
		   & $(\frac{h}{32}, \frac{w}{32})$ 
		& 463 & 37.70 & 27.40 & 299.70    \\
		\bottomrule
	\end{tabular}
	\vspace{10pt}
	\caption{
		\textbf{Influence of the patch and stride size.} In the combination of ``RBase + PTL'', we set the patch and stride in our PTL into different sizes and validate the effectiveness.
	}
	\label{tab:Patch}
	\vspace{-30pt}
\end{table}
With the patch size of $(\frac{h}{8}, \frac{w}{8})$, our performance improves 0.1 dB and 11.43 in terms of PSNR and fMSE when the stride size halves.
The performance still improves 0.08 dB and 2.85 when the stride size reduces again. 
Compared with non-overlapping patches, overlapping patches help capture more accurate content and more plentiful appearance information for match and translation.

Note that our PTL is built on the high resolution features ($128\times128$), 
but the memory costs are comparable with our LTL.
It is benefit from cutting the background into blocks.

\noindent\textbf{Comparison with Self-Attention.}
It is know that Self-Attention calculates the similarity map, 
which is also introduced in our method.
However, self-attention treats the foreground and background equally,
as a result that foreground locations pay high attention to themselves rather than the related backgrounds;
Meanwhile, the high computational costs makes the self-attention unable to work on high-resolution features.
\tabref{tab:ABL} reveals that both our LTL and PTL bring higher performance improvement than self-attention.

\noindent\textbf{Comparison with AdaIN \cite{huang2017arbitrary}. }
Inspired from AdaIN \cite{huang2017arbitrary}, both image harmonization method RainNet \cite{ling2021region} and our PTL use the feature statistics (mean and standard deviation) to represent the appearance.
However, RainNet \cite{ling2021region} applies the appearance of global background to the global foreground,
while our PTL adaptively adjusts each foreground location by content-related background regions.
To prove the effectiveness of our PTL, we replace it with the RAIN module,
which is the key design in RainNet \cite{ling2021region}.
In \tabref{tab:ABL}, we use ``RBase + PTL'' and ``RBase + AdaIN'' to denote the two strategies.
Compared with ``RBase + AdaIN'', adding PTL to our residual baseline brings 0.33dB, 1.08 and 17.69 performance improvement in terms of PSNR, MSE and fMSE.
Both using feature statistics, the advancement of our PTL demonstrates the effectiveness of matching content-related background regions.

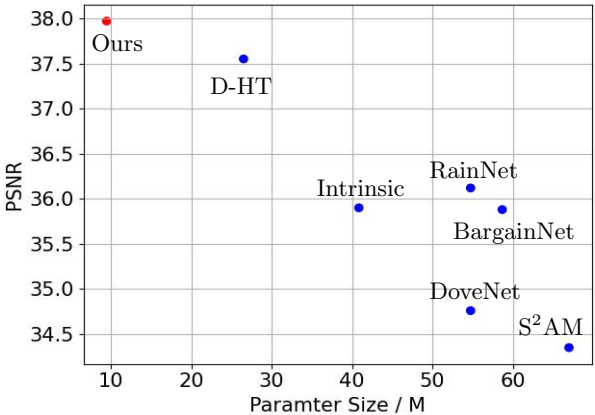
\begin{figure}[t!]
  \centering
  \begin{overpic}[width=0.65\linewidth]{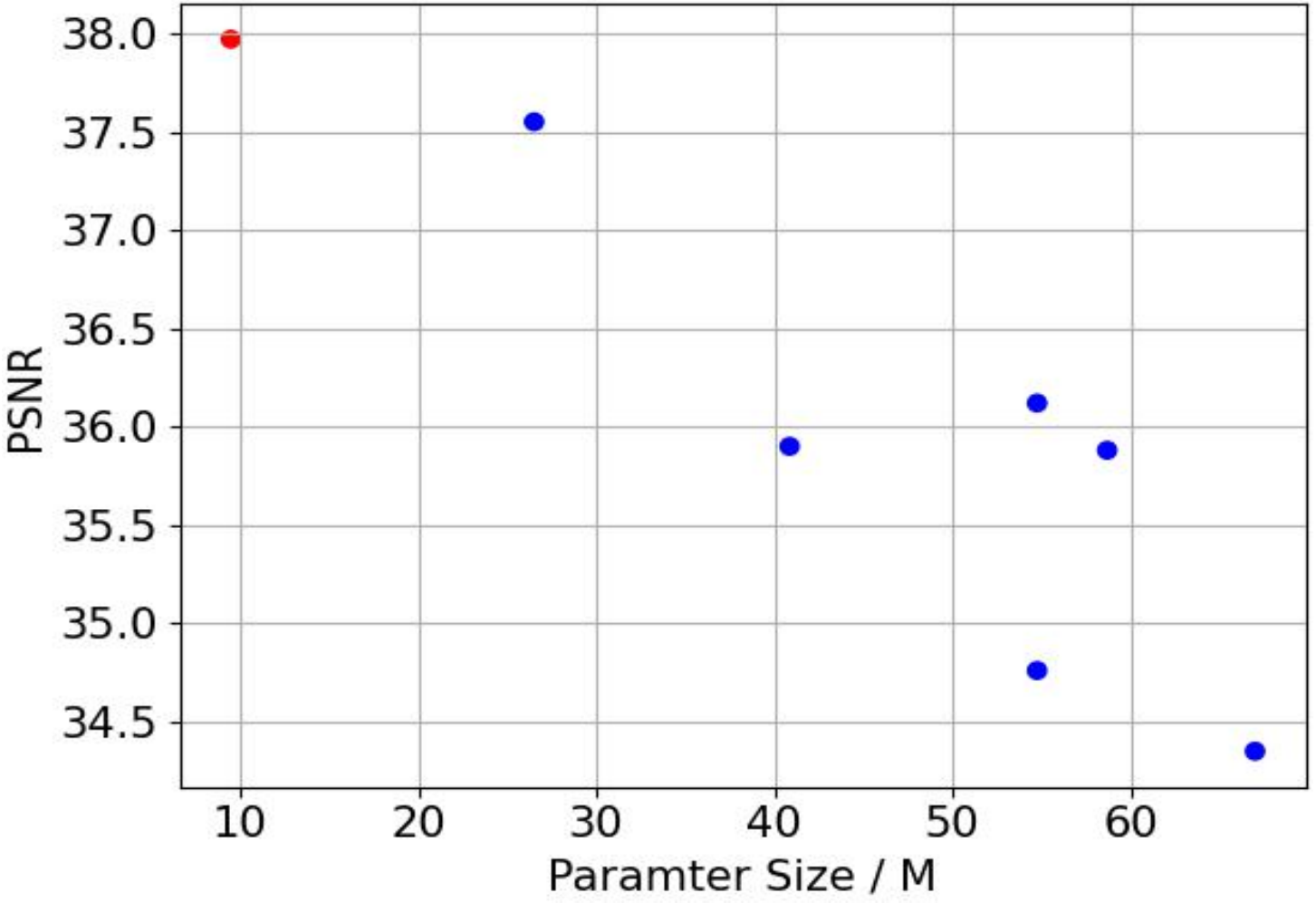}
  \put(15, 60.5){Ours}
  \put(35, 53.5){D-HT}
  \put(53, 36.5){Intrinsic}
  \put(72, 39.5){RainNet}
  \put(76, 29.5){BargainNet}
  \put(72, 19.5){DoveNet}
  \put(87, 13.5){S$^2$AM}
   
  \end{overpic}
  \caption{ \textbf{Parameter size and performance comparison} between our method and six recent methods.
  }
  \label{fig:para}
  \vspace{-10pt}
\end{figure}

\noindent\textbf{Influence of Parameters Size. }
Recent image harmonization methods \cite{Cong2020DoveNet, ling2021region, BargainNet2021, ling2021region, cun2020improving } typically employ the U-net structure as their generators, but their parameter sizes are different because of the channel number and designed modules.
In \figref{fig:para}, we compared our method with five recent methods in terms of performance and parameters size.
With fewer parameters, our method achieves obviously higher performance, which is 0.42 dB higher than the second-best method.

Meanwhile, in \tabref{tab:Parameter} we expand our channel number under the restriction of 1024.
When we double our channel number, the performance of our model increases 0.38 dB, 1.92, 21.35 in terms of PSNR, MSE and fMSE.
This improvement is more significant than the effectiveness of any other strategies. 
However, the parameter number has also increased to about four times the original.
When the number of channels increases to 4 times the original, our performances increase 0.60dB, 3.73, 34.57 in terms of PSNR, MSE and fMSE.
We denote it as ``Ours-p'' in \tabref{tab:quantitative}.
The parameters size of ``Ours-p'' is 117.42 M, which is much larger than normal image harmonization methods.
But it can still be trained with the batch size of 4 on an Nvidia 2080Ti GPU.


\begin{table}[t!]
	\centering
	\renewcommand{\arraystretch}{1.25}
	\renewcommand{\tabcolsep}{1.2mm}
	\begin{tabular}{c|c|ccc}
		\hline\toprule
		\multirow{2}{*}{Channels} & \multirow{2}{*}{Parameters}    & \multicolumn{3}{c}{\textit{iHarmony4}~\cite{Cong2020DoveNet}}     \\
		       &  &  PSNR$\uparrow$ & MSE$\downarrow$ & fMSE$\downarrow$   \\
		\hline
        $C$ &  9.48M          
		& 37.97 & 25.16 & 282.69    \\
	     $C \times 2$  & 37.81M            
	    & 38.35 & 23.24 & 261.34    \\
		 $C \times 3$  & 72.39M  
		& 38.45 & 22.83 & 256.69    \\     
		$C \times 4$ & 117.42M          
		& 38.57 & 21.43 & 248.12    \\
		\bottomrule
	\end{tabular}
	\vspace{10pt}
	\caption{
		\textbf{Influence of the parameters size.} We change the the channel number in our U-net structure and illustrate the impact of increasing parameter size. ``C'' denotes original channel number.
	}
	\label{tab:Parameter}
\end{table}

\section{Conclusion and Limitation}
In this paper, we observe that current image harmonization methods suffer from a performance bottleneck because of not paying due attention to the significant appearance variance in different regions.
To overcome this shortcoming,
we focus on harmonizing the appearance of each foreground location by consulting its related background regions.
Correspondingly, two appearance translation strategies (locations-to-location translation and patches-to-location translation) are designed to adjust the foreground in a coarse-to-fine manner.
Further, we design a reconstruction strategy, which uses the predicted residual to change the foreground appearance.
Extensive visualization results and experimental analysis demonstrate our contributions.
Finally, our method achieves \sArt results on the benchmark dataset \textit{iHarmony4}.
Nevertheless, 
there are still some limitations.
For example,
disharmony patterns in training set are artificially generated. 
Better data generation or better training methods to make the model more robust to the real world is worthy of further exploration.

\clearpage
%
%
\bibliographystyle{splncs04}
\bibliography{eccv}
\end{document}